\documentclass[10pt,twocolumn,letterpaper]{article}

\usepackage{iccv}
\usepackage{times}
\usepackage{epsfig}
\usepackage{graphicx}
\usepackage{amsmath}
\usepackage{amssymb}

\usepackage{enumitem}
\usepackage{placeins}
\usepackage{graphicx}
\usepackage{url}
\usepackage{booktabs}

\usepackage[pagebackref=true,breaklinks=true,letterpaper=true,colorlinks,bookmarks=false]{hyperref}

\iccvfinalcopy 

\def\iccvPaperID{2450} 

\usepackage{subfig}
\usepackage{color}
\newcommand*\bigcdot{\mathpalette\bigcdot@{.5}}
\ificcvfinal\fi

\DeclareMathOperator{\E}{\mathbb{E}}
\begin{document}

\def\mA{\mathcal{A}}
\def\mB{\mathcal{B}}
\def\mC{\mathcal{C}}
\def\mD{\mathcal{D}}
\def\mE{\mathcal{E}}
\def\mF{\mathcal{F}}
\def\mG{\mathcal{G}}
\def\mH{\mathcal{H}}
\def\mI{\mathcal{I}}
\def\mJ{\mathcal{J}}
\def\mK{\mathcal{K}}
\def\mL{\mathcal{L}}
\def\mM{\mathcal{M}}
\def\mN{\mathcal{N}}
\def\mO{\mathcal{O}}
\def\mP{\mathcal{P}}
\def\mQ{\mathcal{Q}}
\def\mR{\mathcal{R}}
\def\mS{\mathcal{S}}
\def\mT{\mathcal{T}}
\def\mU{\mathcal{U}}
\def\mV{\mathcal{V}}
\def\mW{\mathcal{W}}
\def\mX{\mathcal{X}}
\def\mY{\mathcal{Y}}
\def\mZ{\mathcal{Z}}

\def\1n{\mathbf{1}_n}
\def\0{\mathbf{0}}
\def\1{\mathbf{1}}

\def\A{{\bf A}}
\def\B{{\bf B}}
\def\C{{\bf C}}
\def\D{{\bf D}}
\def\E{{\bf E}}
\def\F{{\bf F}}
\def\G{{\bf G}}
\def\H{{\bf H}}
\def\I{{\bf I}}
\def\J{{\bf J}}
\def\K{{\bf K}}
\def\L{{\bf L}}
\def\M{{\bf M}}
\def\N{{\bf N}}
\def\O{{\bf O}}
\def\P{{\bf P}}
\def\Q{{\bf Q}}
\def\R{{\bf R}}
\def\S{{\bf S}}
\def\T{{\bf T}}
\def\U{{\bf U}}
\def\V{{\bf V}}
\def\W{{\bf W}}
\def\X{{\bf X}}
\def\Y{{\bf Y}}
\def\Z{{\bf Z}}

\def\a{{\bf a}}
\def\b{{\bf b}}
\def\c{{\bf c}}
\def\d{{\bf d}}
\def\e{{\bf e}}
\def\f{{\bf f}}
\def\g{{\bf g}}
\def\h{{\bf h}}
\def\i{{\bf i}}
\def\j{{\bf j}}
\def\k{{\bf k}}
\def\l{{\bf l}}
\def\m{{\bf m}}
\def\n{{\bf n}}
\def\o{{\bf o}}
\def\p{{\bf p}}
\def\q{{\bf q}}
\def\r{{\bf r}}
\def\s{{\bf s}}
\def\t{{\bf t}}
\def\u{{\bf u}}
\def\v{{\bf v}}
\def\w{{\bf w}}
\def\x{{\bf x}}
\def\y{{\bf y}}
\def\z{{\bf z}}

\def\balpha{\mbox{\boldmath{$\alpha$}}}
\def\bbeta{\mbox{\boldmath{$\beta$}}}
\def\bdelta{\mbox{\boldmath{$\delta$}}}
\def\bgamma{\mbox{\boldmath{$\gamma$}}}
\def\blambda{\mbox{\boldmath{$\lambda$}}}
\def\bsigma{\mbox{\boldmath{$\sigma$}}}
\def\btheta{\mbox{\boldmath{$\theta$}}}
\def\bomega{\mbox{\boldmath{$\omega$}}}
\def\bxi{\mbox{\boldmath{$\xi$}}}
\def\bnu{\mbox{\boldmath{$\nu$}}}                                  
\def\bphi{\mbox{\boldmath{$\phi$}}}

\def\bDelta{\mbox{\boldmath{$\Delta$}}}
\def\bOmega{\mbox{\boldmath{$\Omega$}}}
\def\bPhi{\mbox{\boldmath{$\Phi$}}}
\def\bLambda{\mbox{\boldmath{$\Lambda$}}}
\def\bSigma{\mbox{\boldmath{$\Sigma$}}}
\def\bGamma{\mbox{\boldmath{$\Gamma$}}}

\newcommand{\myminimum}[1]{\mathop{\textrm{minimum}}_{#1}}
\newcommand{\mymaximum}[1]{\mathop{\textrm{maximum}}_{#1}}    
\newcommand{\mymean}[1]{\mathop{\textrm{mean}}_{#1}}
\newcommand{\myvar}[1]{\mathop{\textrm{Variance}}_{#1}}
\newcommand{\mymin}[1]{\mathop{\textrm{minimize}}_{#1}}
\newcommand{\mymax}[1]{\mathop{\textrm{maximize}}_{#1}}
\newcommand{\mymins}[1]{\mathop{\textrm{min.}}_{#1}}
\newcommand{\mymaxs}[1]{\mathop{\textrm{max.}}_{#1}}  
\newcommand{\myargmin}[1]{\mathop{\textrm{argmin}}_{#1}} 
\newcommand{\myargmax}[1]{\mathop{\textrm{argmax}}_{#1}} 
\newcommand{\myst}{\textrm{s.t. }}

\newcommand{\denselist}{\itemsep -1pt}
\newcommand{\sparselist}{\itemsep 1pt}

\definecolor{pink}{rgb}{0.9,0.5,0.5}
\definecolor{purple}{rgb}{0.5, 0.4, 0.8}   
\definecolor{gray}{rgb}{0.3, 0.3, 0.3}
\definecolor{mygreen}{rgb}{0.2, 0.6, 0.2}

\newcommand{\cyan}[1]{\textcolor{cyan}{#1}}
\newcommand{\red}[1]{\textcolor{red}{#1}}  
\newcommand{\blue}[1]{\textcolor{blue}{#1}}
\newcommand{\magenta}[1]{\textcolor{magenta}{#1}}
\newcommand{\pink}[1]{\textcolor{pink}{#1}}
\newcommand{\green}[1]{\textcolor{green}{#1}} 
\newcommand{\gray}[1]{\textcolor{gray}{#1}}    
\newcommand{\mygreen}[1]{\textcolor{mygreen}{#1}}    
\newcommand{\purple}[1]{\textcolor{purple}{#1}}       

\definecolor{greena}{rgb}{0.4, 0.5, 0.1}
\newcommand{\greena}[1]{\textcolor{greena}{#1}}

\definecolor{bluea}{rgb}{0, 0.4, 0.6}
\newcommand{\bluea}[1]{\textcolor{bluea}{#1}}
\definecolor{reda}{rgb}{0.6, 0.2, 0.1}
\newcommand{\reda}[1]{\textcolor{reda}{#1}}

\newcommand{\mtodo}[1]{{\color{red}$\blacksquare$\textbf{[TODO: #1]}}}
\newcommand{\myheading}[1]{\vspace{1ex}\noindent \textbf{#1}}

\def\changemargin#1#2{\list{}{\rightmargin#2\leftmargin#1}\item[]}
\let\endchangemargin=\endlist
                                               
\newcommand{\cm}[1]{}

\def\xbi{\overline{\x}_i}
\def\wbi{\overline{\w}_{(i)}}
\def\wb{\overline{\w}}
\def\Ib{\overline{\I}}
\def\invC{\C^{-1}}
\def\invCi{\C_{(i)}^{-1}}
\def\ab{\overline{\balpha}}
\def\abi{\overline{\balpha}_{(i)}}
\def\Kb{\overline{\K}}
\def\Xb{\overline{\X}}
\def\kbi{\overline{\k}_{i}}
\def\Kzz{\K_{\z\z}}
\def\Kzx{\K_{\z\x}}
\def\Xsub{\X_{sub}}
\def\ssub{\s_{sub}}
\def\wbsub{\overline{\w}_{sub}}
\def\dsub{\d_{sub}}
\def\invCsub{\C^{-1}_{sub}}
\def\etal{\emph{et al}.}
\def\etals{\emph{et al}. }
\def\DS{\textcolor{red}}
\newcommand{\norm}[1]{\left\lVert#1\right\rVert}

\def\subFigSzab{\linewidth}
\title{Shadow Removal via Shadow Image Decomposition}

\author{Hieu Le\\
Stony Brook University\\
New York, 11794, USA\\
{\tt\small hle@cs.stonybrook.edu}
\and
Dimitris Samaras\\
Stony Brook University\\
New York, 11794, USA\\
{\tt\small samaras@cs.stonybrook.edu}
}

\maketitle

\begin{abstract}
We propose a novel deep learning method for shadow removal. Inspired by physical models of shadow formation, we use a linear illumination transformation to model the shadow effects in the image that allows the shadow image to be expressed as a combination of the shadow-free image, the shadow parameters, and a matte layer.  We use two deep networks, namely SP-Net and M-Net, to predict the shadow parameters and the shadow matte respectively. This system allows us to remove the shadow effects on the images. We train and test our framework on the most challenging shadow removal dataset (ISTD). Compared to the state-of-the-art method, our model achieves a 40\% error reduction in  terms of root mean square error (RMSE) for the shadow area, reducing RMSE from 13.3 to 7.9. Moreover, we create an augmented ISTD dataset based on an image decomposition system by modifying the shadow parameters to generate new synthetic shadow images. Training our model on this new augmented ISTD dataset further lowers the RMSE on the shadow area to 7.4. 
\end{abstract}


\section{Introduction}
Shadows are cast whenever a light source is blocked by an object. Shadows often confound computer vision algorithms such as segmentation, tracking, or recognition. The appearance of shadow edges is hard to distinguish from edges due to material changes~\cite{shadow_edge}. Dark albedo material regions can be easily misclassified as shadows~\cite{m_Le-etal-ECCV18}. Thus many methods have been proposed to identify and remove  shadows from images.

\begin{figure}[h]
    \centering
    \includegraphics[width=0.5\textwidth]{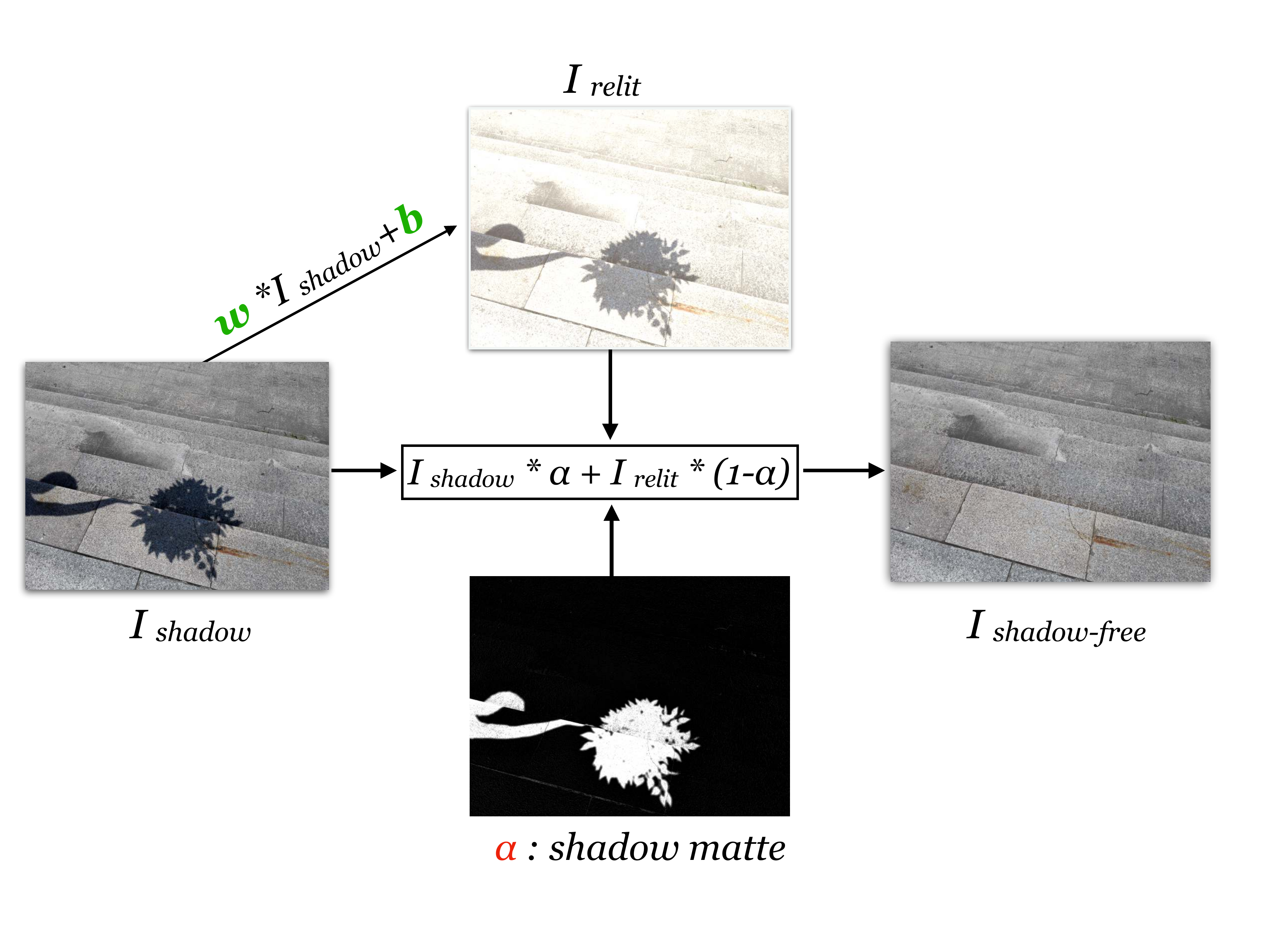}
    \caption{\textbf{Shadow Removal via Shadow Image Decomposition.} A shadow-free image $I_{\textrm{shadow-free}}$ can be expressed in terms of a shadow image $I_{\textrm{shadow}}$, a relit image $I_{\textrm{relit}}$ and a shadow matte $\alpha$. The relit image is a linear transformation of the shadow image. The two unknown factors of this system are the shadow parameters $(w,b)$ and the shadow matte layer $\alpha$. We use two deep networks to estimate these two unknown factors. 
    }
    \label{fig:Teaser}
\end{figure}
Early shadow removal work was based on  physical shadow models \cite{Barrow1978}. A common approach is to formulate the shadow removal problem using an image formation model, in which the image is expressed in terms of material properties and a light source-occluder system that casts shadows. Hence, a shadow-free image can be obtained by estimating the parameters of the source-occluder system and then reversing the shadow effects on the image~\cite{Finlayson06,huang11,guoPami,Shor08}. These methods relight the shadows in a physically plausible manner.
However, estimating the correct solution for such illumination models is non-trivial and requires considerable processing time or user assistance\cite{Zhang15,Chuang2003}.

 On the other hand, recently published large-scale datasets \cite{Qu_2017_CVPR,Wang_2018_CVPR,Vicente-etal-ECCV16} allow the use of deep learning methods for shadow removal. In these cases, a network is trained in an end-to-end fashion to map the input shadow image to a shadow-free image. The success of these approaches shows that deep networks can effectively learn transformations that relight shadowed pixels. However, the actual physical properties of shadows are ignored, and there is no guarantee that the networks would learn physically plausible transformations. Moreover, there are still well known issues with images generated by deep networks: results tend to be blurry \cite{isola2017image,zhang2016colorful} and/or contain artifacts \cite{odena2016deconvolution}. How to improve the quality of generated images is an active research topic~\cite{karras2018progressive,wang2018pix2pixHD}.

In this work, we propose a novel method for shadow removal that takes advantage of both shadow illumination modelling and deep learning. Following early shadow removal works, we propose to use a simplified physical illumination model to define the mapping between shadow pixels and their shadow-free counterparts.
 
 Our proposed illumination model is a linear transformation consisting of a scaling factor and an additive constant - per color channel - for the whole umbra area of the shadow. These scaling factors and additive constants are the parameters of the model, see Fig. \ref{fig:Teaser}. 
 The illumination model plays a key role in our method: with correct parameter estimates, we can use the model to remove shadows from images. We propose to train a deep network (SP-Net) to automatically estimate the parameters of the shadow model. Through training, SP-Net learns a mapping function from input shadow images to illumination model parameters. 
 
 


Furthermore, we use a shadow matting technique \cite{Chuang2003,guoPami,Zhang15} to handle the penumbra area of the shadows. We incorporate our illumination model into an image decomposition formulation~\cite{Porter1984,Chuang2003}, where the shadow-free image is expressed as a combination of the shadow image, the parameters of the shadow model, and a shadow density matte. This image decomposition formulation allows us to reconstruct the shadow-free image, as illustrated in Fig. \ref{fig:Teaser}. The shadow parameters $(w,b)$ represent the transformation from the shadowed pixels to the illuminated pixels. The shadow matte represents the per-pixel linear combination of the relit image and the shadow image, which results to the shadow-free image. 
Previous work often requires user assistance\cite{Gong16} or solving an optimization system \cite{Levin08} to obtain the shadow mattes. In contrast, we propose to train a second network (M-Net) to accurately predict shadow mattes in a fully automated manner. 


We train and test our proposed SP-Net and M-Net on the ISTD dataset \cite{Wang_2018_CVPR}, which is the largest and most challenging available dataset for shadow removal. SP-Net alone (no matting) outperforms the state-of-the-art \cite{Gong16} in shadow removal by 29\% in terms of RMSE on shadow areas, from 13.3 to 9.5 RMSE. Our full system with both SP-Net and M-Net further improves the overall results by another 17\%, which yields a RMSE of 7.9.

Our proposed method can realistically modify the shadow effects in the images. First we estimate the shadow parameters and shadow matte from an image. We then add the shadows back into the shadow-free image with a set of modified shadow parameters. As we change the parameters, the shadow effects change accordingly. In this manner, we can synthetize additional shadow images that serve as augmented training data. Training our system on ISTD plus our newly synthesized images further lowers the RMSE on the shadow areas by 6\%, compared to our model trained on the original ISTD dataset.

The main contributions of this work are:
\begin{itemize}
    \item We propose a new deep learning approach for shadow removal, grounded by a simplified physical illumination model and an image decomposition formulation.
    \item We propose a method for shadow image augmentation based on our simplified physical illumination model and the image decomposition formulation.
    \item Our proposed method achieves state-of-the-art shadow removal results on the ISTD dataset.
\end{itemize}

The pre-trained model, shadow removal results, and more details can be found at: \url{www3.cs.stonybrook.edu/~cvl/projects/SID/index.html}

\section{Related Works}
\label{sec:related}
\textbf{Shadow Illumination Models:}
Early research on shadow removal is motivated by physical modelling of illumination and color \cite{Finlayson06,Finlayson01,Finlayson02,Drew03recoveryof}.
Barrow \& Tenenbaum \cite{Barrow1978} define an intrinsic image algorithm that separates images into the intrinsic components of reflectance and shading.  Guo \etal~\cite{guoPami} simplify this model to represent the relationship between the shadow pixels and shadow-free pixels via a linear system. 
They estimate the unknown factors via pairing shadow and shadow-free regions. Similarly, Shor \& Lischinki \cite{Shor08} propose an illumination model for shadows in which there is an affine relationship between the lit and shadow intensities at a pixel, including 4 unknown parameters. They define two strips of pixels: one in the shadowed area and one in the lit area to estimate their parameters.
Finlayson \etal \cite{Finlayson09} create an illuminant-invariant
image for shadow detection and removal. Their work is based on an insight that the shadowed pixels differ from their lit pixels by a scaling factor. Vicente \etal~\cite{Vicente-etal-PAMI18,vicentesingle}  propose a method for shadow removal where they suggest that the color of the lit region can be transferred to the shadowed region via histogram equalization. 


\textbf{Shadow Matting:} 
Matting, introduced by Porter \& Duff \cite{Porter1984}, is an effective tool to handle soft shadows. However, it is non-trivial to compute the shadow matte from a single image.
Chuang \etal~\cite{Chuang2003} use image matting for shadow editing to transfer the shadows between different scenes. They compute the shadow matte from a sequence of frames in a video captured from a static camera.  
Guo et al. \cite{guoPami} and Zhang \etal~\cite{Zhang15} both use a shadow matte for their shadow removal frameworks, where they estimate the shadow matte via the closed-form solution of Levin \etal~\cite{Levin08}. 

\textbf{Deep-Learning Based Shadow Removal:} 
Recently published large-scale datasets~\cite{Vicente-etal-ECCV16, Wang_2018_CVPR,Qu_2017_CVPR} enable training deep-learning networks for shadow removal.
The Deshadow-Net of Qu \etal~\cite{Qu_2017_CVPR} is trained to remove  shadows in an end-to-end manner.  Their network extracts multi-context features across different layers of a deep network to predict a shadow matte. This shadow matte is different from ours as it contains both the density and color offset of the shadows. The ST-CGAN proposed by Wang \etal~\cite{Wang_2018_CVPR} for both shadow detection and removal is a conditional GAN-based framework \cite{isola2017image} for shadow detection and removal. Their framework is trained to predict the shadow mask and shadow-free image in an unified manner, they use GAN losses to improve performance. 

Inspired by early work, our framework outputs the shadow-free image based on a physically inspired shadow illumination model and a shadow matte. We, however, estimate the parameters of our model and the shadow matte via two deep networks in a fully automated manner.

\begin{figure*}[!ht]
 \centering
    \includegraphics[width=\textwidth]{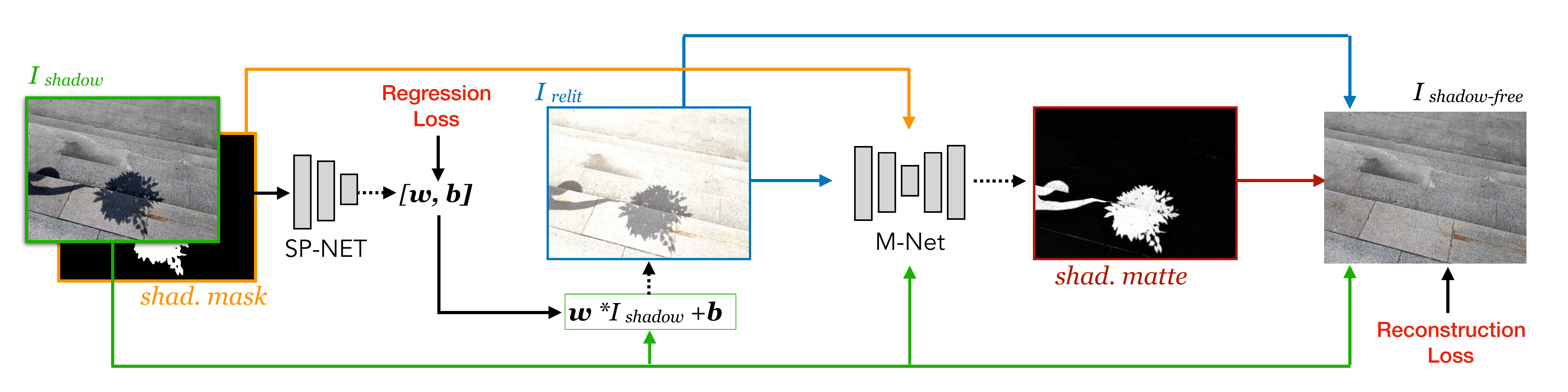}
    \caption{\textbf{Shadow Removal Framework.} The shadow parameter estimator network SP-Net takes as input the shadow image and the shadow mask to predict the shadow parameters $(w,b)$. The relit image  $I^{\textrm{relit}}$ is then computed via Eq. \ref{eq:relit} using the estimated parameters from SP-Net. The relit image, together with the input shadow image and the shadow mask are then input into the shadow matte prediction network M-Net to get the shadow matte layer $\alpha$. The system outputs the shadow-free image via Eq. \ref{eq:decom}, using the shadow image, the relit image, and the shadow matte. SP-Net learns to predict the shadow parameters $(w,b)$, denoted as the regression loss. M-Net learns to minimize the $L_1$ distance between the output of the system and the shadow-free image (reconstruction loss).
    }
    \label{fig:framework}
\end{figure*}

\section{Shadow and Image Decomposition Model}
\label{sec:decom_model}
\subsection{Shadow Illumination Model}
Let us begin by describing our shadow illumination model. We aim to find a mapping function $T$ to transform a shadow pixel $I^{\textrm{shadow}}_x$ to its non-shadow counterpart: $I_x^{\textrm{shadow-free}} = T(I^{\textrm{shadow}}_x,w)$ where $w$ are the parameters of the model. 
The form of $T$ has been studied in depth in  previous work as  discussed in Sec. \ref{sec:related}. 


In this paper, similar to the model of Shor \& Lischinski \cite{Shor08}, we use a linear function to model the relationship between the lit and shadowed pixels. The intensity of a lit pixel is formulated as:
\begin{equation}
\label{eq:shadfree}
 I^{\textrm{shadow-free}}_x(\lambda) =  L^d_x(\lambda)R_x(\lambda) +   L^a_x(\lambda)R_x(\lambda)
\end{equation}
where $ I^{\textrm{shadow-free}}_x(\lambda)$ is the intensity reflected from point $x$ in the scene at wavelength $\lambda$, $L$ and $R$ are the illumination and reflectance respectively, $L^d$ is the direct illumination and $L^a$ is the ambient illumination.

To  cast a shadow on point $x$,  an occluder blocks the direct illumination and a portion of the ambient illumination that would otherwise arrive at $x$. The shadowed intensity at $x$ is:

\begin{equation}
\label{eq:shad}
 I^{\textrm{shadow}}_x(\lambda) =  a_x(\lambda)L^a_x(\lambda)R_x(\lambda)
\end{equation}
where $a_x(\lambda)$ is the attenuation factor indicating the remaining fraction of the ambient illumination that arrives at point $x$ at wavelength $\lambda$.
Note that Shor \& Lischinski further assume that $a_x(\lambda)$ is the same for all wavelengths $\lambda$ to simplify their model. This assumption implies that the environment light has the same color from all directions.

From Eq.\ref{eq:shadfree} and \ref{eq:shad}, we can express the shadow-free pixel as a linear function of the shadowed pixel:
\begin{equation}
\label{eq:maineq}
 I^{\textrm{shadow-free}}_x(\lambda) =  L^d_x(\lambda)R_x(\lambda) +  a_x(\lambda)^{-1}I^{\textrm{shadow}}_x(\lambda)
\end{equation}

We assume that this linear relation is preserved throughout the color acquisition process of the camera \cite{Finlayson16}. Therefore, we can express the color intensity of the lit pixel $x$ as a linear function of its shadowed value:
\begin{equation}
\label{eq:trans}
    I_x^{\textrm{shadow-free}}(k) = w_k \times I_x^{\textrm{shadow}}(k) + b_k
\end{equation}
where $I_x(k)$ represents the value of the pixel $x$ on the image $I$ in  color channel $k$ ($k\in {\textrm{R,G,B color channel}}$), $b_k$ is the response of the camera to  direct illumination, and $w_k$ is responsible for the attenuation factor of the ambient illumination at this pixel in this color channel. We model each color channel independently to account for possibly different spectral characteristics of the material in shadow as well as the sensor.

We further assume that the two vectors $w =[w_R,w_G,w_B]$ and $b=[b_R,b_G,b_B]$ are constant across all pixels $x$ in the umbra area of the shadow. Under this assumption, we can easily estimate the values of $w$ and $b$  given the shadow and shadow-free image using linear regression.  We refer to $(w,b)$ as the \textit{shadow parameters} in the rest of the paper. 

In Sec. \ref{sec:framework}, we show that we can train a deep-network to estimate these vectors from a single image.




\subsection{Shadow Image Decomposition System}
\label{sec:decom}
We plug our proposed shadow illumination model into the following well-known image decomposition system \cite{Chuang2003,Porter1984,Smith1996,Wright}. The system models the shadow-free image using the shadow image, the shadow parameter, and the shadow matte. The shadow-free image can be expressed as:

\begin{equation}
    I^{\textrm{shadow-free}} = I^{\textrm{shadow}} \cdot \alpha + I^{\textrm{relit}}\cdot (1-\alpha)
    \label{eq:decom}
\end{equation}
where  $I^{\textrm{shadow}}$ and $I^{\textrm{shadow-free}}$ are the shadow and shadow-free image respectively, $\alpha$ is the matting layer, and $I^{\textrm{relit}}$ is the relit image. We define $\alpha$ and $I^{\textrm{relit}}$ below.

Each pixel $i$ of the relit image  $I^{\textrm{relit}}$  is computed by:

\begin{equation}
    I_i^{\textrm{relit}} = w \cdot I_i^{\textrm{shadow}} +b
    \label{eq:relit}
\end{equation}
which is the shadow image  transformed by the illumination model of Eq. \ref{eq:trans}. This transformation maps the shadowed pixels to their shadow-free values.

The matting layer $\alpha$ represents the per-pixel coefficients of the linear combination of the relit image and the input shadow image that results into the shadow-free image. Ideally, the value of $\alpha$ should be 1 at the non-shadow area and 0 at the umbra of the shadow area. For the pixels in the penumbra of the shadow, the value of $\alpha$ gradually changes near the shadow boundary.

The value of $\alpha$ at pixel $i$ based on  the shadow image, shadow-free image, and relit image, follows from Eq. \ref{eq:decom} :
\begin{equation}
\label{eq:alpha}
\alpha_i=\frac{{I_i}^{\textrm{shadow-free}}-{I_i}^{\textrm{relit}}}{{I_i}^{\textrm{shadow}}-{I_i}^{\textrm{relit}}}
\end{equation}

We use the image decomposition of Eq. \ref{eq:decom} for our shadow removal framework. The  unknown factors are the shadow parameters $(w,b)$ and the shadow matte $\alpha$. We present our method that uses two deep networks, SP-Net and M-Net, to predict these two factors in the following section. In Sec.\ref{sec:aug}, we propose a simple method to modify the shadows for an image in order to  augment the training data.


\section{Shadow Removal Framework}
\label{sec:framework}
Fig. \ref{fig:framework} summarizes our framework. The shadow parameter estimator network SP-Net takes as input the shadow image and the shadow mask to predict the shadow parameters $(w,b)$. The relit image  $I^{\textrm{relit}}$ is then computed via Eq. \ref{eq:relit} with the estimated parameters from SP-Net. The relit image, together with the input shadow image and the shadow mask is then input into the shadow matte prediction network M-Net to get the shadow matte $\alpha$. The system outputs the shadow-free image via Eq. \ref{eq:decom}.

\subsection{Shadow Parameter Estimator Network}
\label{sec:SP-Net}
In order to recover the illuminated intensity at the shadowed pixel, we need to estimate the parameters of the linear model in Eq. \ref{eq:trans}. Previous work has proposed different methods to estimate the parameters of a shadow illumination model \cite{Shor08,Gong16,guoPami,Finlayson02,Finlayson09,Drew03recoveryof}. 
In this paper, we train SP-Net, a deep network model, to directly predict the shadow parameters from the input shadow image.

To train SP-Net, we first generate training data.
Given a training pair of a shadow image and a shadow-free image, we estimate the parameters of our linear illumination model using a least squares method \cite{cook1986}. For each shadow image, we first erode the shadow mask by 5 pixels in order to define a region that does not contain the partially shadowed (penumbra) pixels. Mapping these shadow pixel values to the corresponding values in the shadow-free image, gives us a linear regression system, from which we calculate $w$ and $b$.  
We compute parameters for each of  the three RGB color channels  and then combine the learned coefficients to form a  6-element vector. This vector is used as the targeted output to train SP-Net. The input for SP-Net is the input shadow image and the associated shadow mask. We train SP-Net to minimize the $L_1$ distance between the output of the network and these computed shadow parameters. 

We develop SP-Net by customizing a ResNeXt \cite{Xie_2017_CVPR} model that is pre-trained on ImageNet \cite{imagenet_cvpr09}. 
Notice that while we use the ground truth shadow mask for training, during testing we estimate shadow masks using the shadow detection network proposed by Zhu \etal \cite{zhu18b}.

\subsection{Shadow Matte Prediction Network}
\label{Sec:MNet}

\def\subboxsize{0.16\textwidth}
\begin{figure*}[t]
    \centering
    \includegraphics[width=\textwidth]{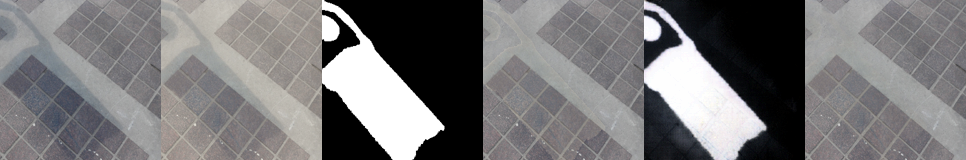}\\
    \makebox[\subboxsize]{Input}
    \makebox[\subboxsize]{Relit}
    \makebox[\subboxsize]{Shad. Mask}
    \makebox[\subboxsize]{Using S.Mask}
    \makebox[\subboxsize]{Shad. Matte}
    \makebox[\subboxsize]{Using S.Matte}

    \caption{\textbf{A comparison of the ground truth shadow mask and our shadow matte.} From the left to right: The input image, the relit image computed from the parameters estimated via SP-Net, the ground truth shadow mask, the final results when we use the shadow mask, the shadow matte computed using our M-Net, and the final shadow-free image  when we use the shadow matte to combine the input and relit image. The matting layer handles the soft shadow and does not generate visible boundaries in the final result. \em{(Please view in magnification on a digital device to see the difference more clearly.)}}
    \label{fig:matte}
\end{figure*}

Our linear illumination model (Eq. \ref{eq:trans}) can relight the pixels in the umbra area (fully shadowed). The shadowed pixels in the penumbra (partially shadowed) region are more challenging as the illumination changes gradually across the shadow boundary \cite{huang11}. A binary shadow mask cannot model this gradual change. Thus, using a binary mask within the decomposition model in Eq. \ref{eq:decom} will generate an image with visible boundary artifacts. A solution for this is shadow matting where the soft shadow effects are expressed via the values of a blending layer.

In this paper, we train a deep network, M-Net, to predict this matting layer. In order to train M-Net, we use Eq. \ref{eq:decom} to compute the output of our framework where the shadow matte is the output of M-Net. Then the loss function that drives the training of M-Net is the $L_1$ distance between output image and ground truth training shadow-free image, marked as ``reconstruction loss'' in Fig. \ref{fig:framework}. This is equivalent to computing the actual value of the shadow matte via Eq. \ref{eq:alpha} and then training M-Net to directly output this value. 

Fig. \ref{fig:matte} illustrates the effectiveness of our shadow matting technique. We show in the figure two shadow removal results which are computed using a ground-truth shadow mask and a shadow matte respectively. This shadow matte is computed by our model. One can see that using the binary shadow mask to form the shadow-free image  creates visible boundary artifacts as it ignores the penumbra. The shadow matte  from our model captures well the soft shadow and generates an image without shadow boundary artifacts.


We design M-Net based on U-Net \cite{unet15a}. The M-Net inputs  are the shadow image, the relit image, and the shadow mask. We use the shadow mask as  input to M-Net since the matting layer can be considered as a relaxed shadow mask where each value represents the strength of the shadow effect at the location rather than just the shadow presence.

\section{Experiments}

\subsection{Dataset and Evaluation Metric}
We train and evaluate on the ISTD dataset \cite{Wang_2018_CVPR}. ISTD consists of image triplets: shadow image, shadow mask, and shadow-free image, captured from different scenes. The training split has 1870 image triplets from 135 scenes, whereas the testing split has 540 triplets from 45 scenes.

\def\subboxsize{0.3\subFigSzab}
\begin{figure}[]
 \centering
    \includegraphics[width=0.45\textwidth]{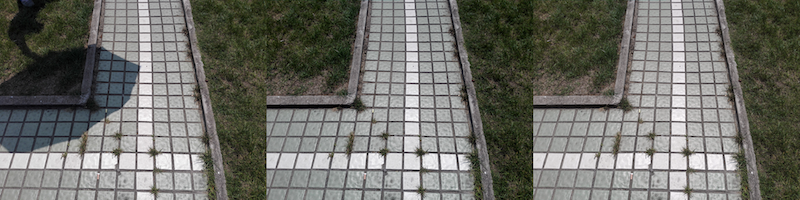}
     \makebox[\subboxsize]{Shad. Image}
    \makebox[\subboxsize]{Original GT}
    \makebox[\subboxsize]{Corrected GT}
    \caption{\textbf{An example of our color correction method.} From left to right: input shadow image, provided shadow-free ground truth image (GT) from ISTD dataset, and the GT image corrected by our method. Comparing to the input shadow image on the non-shadow area only, the root-mean-square distance of the original GT is 12.9. This value on our corrected GT becomes 2.9. 
    }
    \label{fig:fix_dataset}
\end{figure}

We notice that the testing set of the  ISTD dataset needs to be adjusted since the shadow images and the shadow-free images have inconsistent colors. This is a well known issue mentioned in the original paper \cite{Wang_2018_CVPR}. The reason is that the shadow and shadow-free image pairs were captured at different times of the day which resulted in slightly different environment lights for each image.
For example, Fig. \ref{fig:fix_dataset} shows a shadow  and shadow-free image pair. The root-mean-square difference between these two images in the non-shadow area is 12.9. This color inconsistency appears frequently in the testing set of the ISTD dataset. On the whole testing set, the root-mean-square distance between the shadow images and shadow-free images in the non-shadow area is 6.83, as computed by Wang \etal \cite{Wang_2018_CVPR}. 

In order to mitigate this color inconsistency, we use linear regression to transform the pixel values  in the non-shadow area of each shadow-free image to map into  their counterpart values in the shadow image. We use a linear regression for each color-channel, similar to our method for relighting the shadow pixels in Sec. \ref{sec:SP-Net}. This simple transformation transfers the color tone and brightness of the shadow image to its shadow-free counterpart. The third column of Fig. \ref{fig:fix_dataset} illustrates the effect of our color-correction method. Our proposed method reduces the root-mean-square distance between the shadow-free image and the shadow image from 12.9 to 2.9. The error reduction for the whole testing set of ISTD goes from 6.83 to 2.6. 

\subsection{Shadow Removal Evaluation}
 
We evaluate our method on the adjusted testing set of the ISTD dataset. For metric evaluation we follow \cite{Wang_2018_CVPR} and compute the RMSE in the LAB color space on the shadow area, non-shadow area, and the whole image, where all shadow removal results are re-sized into $256\times256$ to compare with the ground truth images at this size.  
Note that in contrast to other methods that only output shadow free images at that resolution, our shadow removal system works for input images of any size. 
Since our method requires shadow masks, we use the model proposed by Zhu \etal \cite{zhu18b} pre-trained on the SBU dataset \cite{Vicente-etal-ECCV16} for detecting shadows. We take the model provided by the author and fine-tune it on the ISTD dataset for 3000 epochs. This model achieves 2.2 Balance Error Rate on the ISTD testing set. To remove the shadow effect in the image, we first use SP-Net to compute the shadow parameters $(w,b)$ using the input image and the shadow mask computed from the shadow detection network. We use $(w,b)$ to compute a relit image which is input to M-Net, together with the input image and the shadow mask to output a matte layer. We obtain the final shadow removal result via Eq. \ref{eq:decom}. 
In Table \ref{tab:basic}, we compare the performance of our method with the recent shadow removal methods of Guo \etal \cite{guoPami}, Yang \etal \cite{Yang12}, Gong \etal \cite{Gong16}, and Wang \etal \cite{Wang_2018_CVPR}. All numbers are computed on the adjusted testing images so that they are directly comparable. The first row shows the numbers for the input shadow images, i.e. no shadow removal performed.


We first evaluate our shadow removal performance using only SP-Net, i.e.  we use the binary shadow mask computed by the shadow detector to form the shadow-free image from the shadow image and the relit image. 
The binary shadow mask is obtained by simply thresholding the output of the shadow detector with a threshold of 0.95. As shown in  column ``\textit{SP-Net}'' (third from the right) in Fig. \ref{fig:main}, SP-Net correctly estimates the shadow parameters to relight the shadow area. Even with visible shadow boundaries, SP-Net alone outperforms the previous state-of-the-art, reducing the RMSE on the shadow area by 29\%, from 13.3 to 9.5.

\def\subboxsize{0.22\subFigSzab}
\begin{figure}[]
 \centering
    \includegraphics[width=0.45\textwidth]{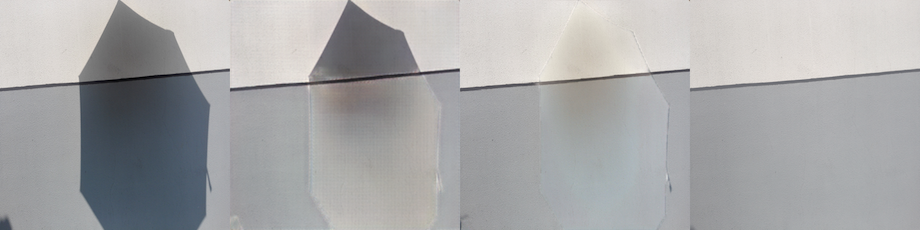}
    \includegraphics[width=0.45\textwidth]{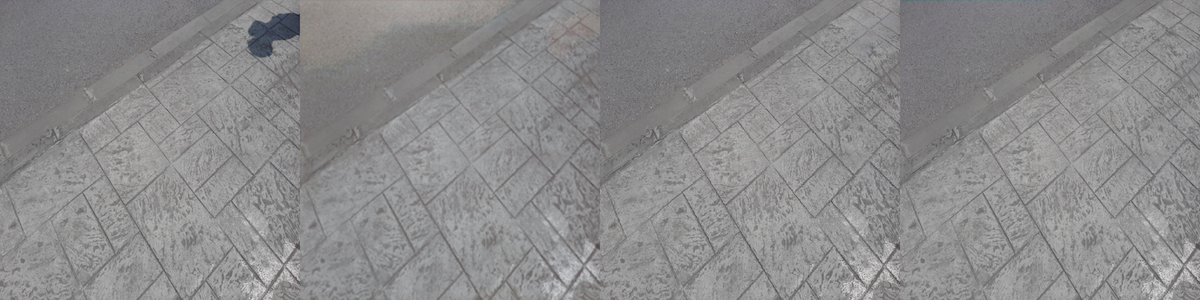}
       \includegraphics[width=0.45\textwidth]{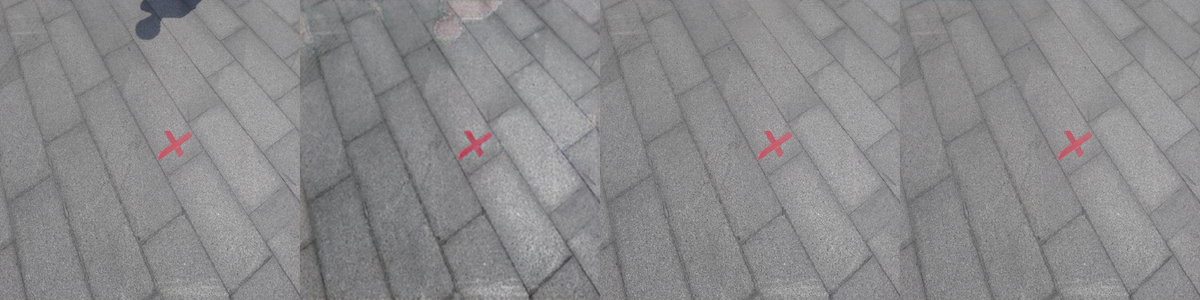}
    \includegraphics[width=0.45\textwidth]{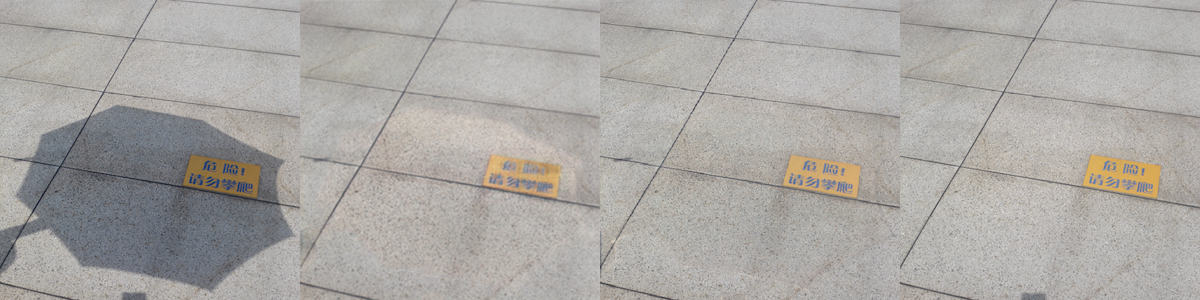}
    \includegraphics[width=0.45\textwidth]{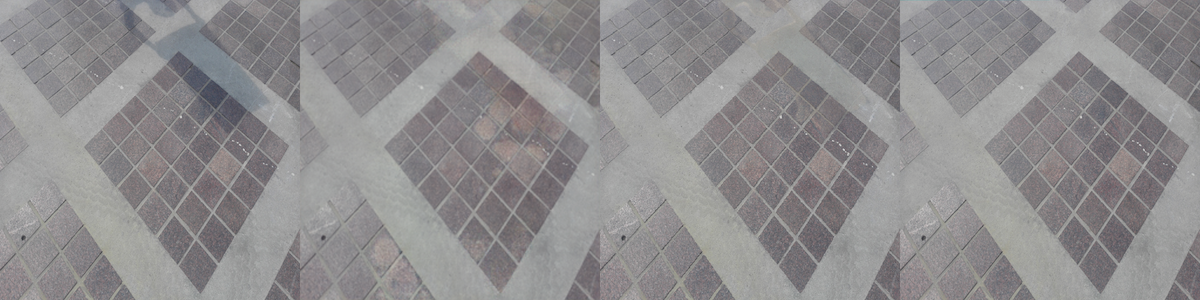} 
     \makebox[\subboxsize]{Input}
    \makebox[\subboxsize]{Wang \etal \cite{Wang_2018_CVPR}}
        \makebox[\subboxsize]{Ours}
    \makebox[\subboxsize]{GT}
    \caption{\textbf{Comparison of shadow removal between our method and ST-CGAN \cite{Wang_2018_CVPR}.} ST-CGAN tends to produce blurry images, random artifacts, and incorrect colors of the lit pixels while our method handles all cases well. 
    }
    \label{fig:stcgan}
\end{figure}

We then evaluate the shadow removal results using both SP-Net and M-Net, denoted as 
``\textit{SP+M-Net}'' in Tab. \ref{tab:basic} and Fig. \ref{fig:main}. 
As shown in Fig. \ref{fig:main}, the results of M-Net do not contain  boundary artifacts. 
In the third row of Fig. \ref{fig:main}, SP-Net overly relights the shadow area but the shadow matte computed from M-Net effectively corrects these errors. This is because M-Net is trained to blend the relit and shadow images  to create the shadow-free image. Therefore, M-Net  learns to output a smaller weight for a pixel that is overly lit by SP-Net.
Using the matte layer of M-Net further reduces the RMSE on the shadow area by 17\%, from 9.5 to 7.9. 

Overall, our method generates better results than other methods.  Our method does a better job at estimating the overall illumination changes compared to the model of Gong \etal, which tends to overly relight  shadow pixels, as shown in Fig. \ref{fig:main}. Our method does not show color inconsistencies within the relit area contrary to all other methods.  Fig. \ref{fig:stcgan}  qualitatively compares  our method and ST-CGAN, which illustrates common issues present in images generated by deep networks \cite{isola2017image,zhang2016colorful}. ST-CGAN generally generates blurry images and introduces random artifacts. Our method, albeit not perfect, handles all cases well.  

Our method fails to recover the shadow-free pixels properly as shown  in Fig. \ref{fig:fail}. The first row, shows how our method overly relights the shadowed area while in the second row, the color of the lit area is incorrect. 

Finally, we trained and evaluated two alternative designs that do not require shadow masks as input: (1) The first is an end-to-end shadow-removal system where we jointly train a shadow detector together with our proposed SP-Net and M-Net. This framework is harder to train due to the increase in the number of network parameters. (2) The second is a version of our framework that does not input the shadow masks into both SP-Net and M-Net. Hence, SP-Net and M-Net need to learn to localize the shadow areas implicitly. As can be seen in the two bottom rows of  Tab. \ref{tab:basic}, both designs achieved slightly worse shadow removal results than our main setting.

\begin{table}[]
\centering
\caption{\textbf{Shadow removal results of our networks compared to state-of-the-art  shadow removal methods on the adjusted ground truth.} $(*)$ The method of Gong \etal \cite{Gong16} is an interactive method that defines the shadow/non-shadow regions via user inputs, thus generates minimal error on the non-shadow area. The metric is RMSE (the lower, the better). Best results are in bold.}
\begin{tabular}{lccc}
\toprule
Methods                    & Shadow& Non-Shadow& All  \\ 
\midrule
Input Image                & 40.2  & 2.6 & 8.5\\ 
\midrule
Yang \etal~\cite{Yang12}                 & 24.7  & 14.4 & 16.0\\ 
Guo \etal~\cite{guoPami}                  & 22.0  & 3.1 & 6.1\\ 
Wang \etal \cite{Wang_2018_CVPR}    & 13.4  & 7.7 & 8.7\\ 
Gong \etal~\cite{Gong16}            & 13.3  & \textbf{2.6*} & 4.2\\ 
\midrule
SP-Net (Ours)  & 9.5  &3.2 &4.1\\ 
SP+M-Net (Ours)   & \textbf{7.9}  &3.1 &\textbf{3.9}\\
\midrule
\midrule
\multicolumn{4}{c}{Our Method with Alternative Settings}\\
\midrule
With a Shad. Detector  & 8.4  &5.0 &5.5\\ 
No Input Shadow Mask   & 8.3  &4.9 &5.4\\
\midrule
\end{tabular}
\label{tab:basic}
\end{table}

\def\subboxsize{0.27\subFigSzab}
\begin{figure}[]
 \centering

    \includegraphics[width=0.45\textwidth]{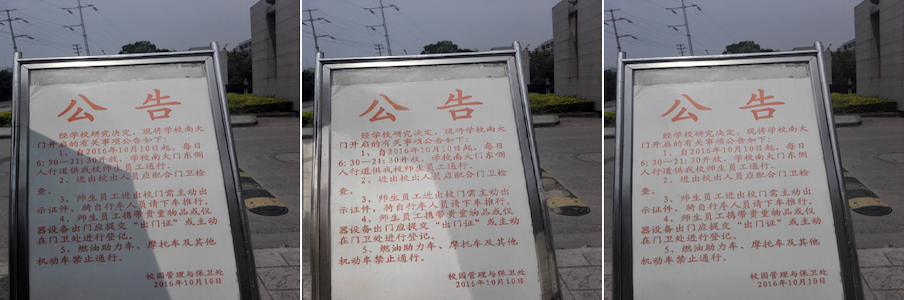}
    \includegraphics[width=0.45\textwidth]{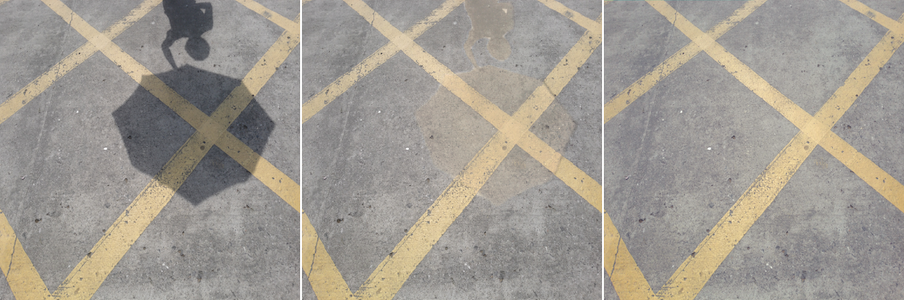}
     \makebox[\subboxsize]{Input}
    \makebox[\subboxsize]{Ours}
        \makebox[\subboxsize]{GT}
 
    \caption{\textbf{Failure cases of our method.} In the first row, our method overly lights up the shadow area. In the second row, our method generates incorrect colors.  
    }
    \label{fig:fail}
\end{figure}

\subsection{Dataset Augmentation via Shadow Editing}
\label{sec:aug}
Many deep learning work focus on learning from more easily obtainable, weakly-supervised, or synthetic data \cite{Buhmann12weak,LeICCV2017,Liu2014FashionPW,Liu2013WeaklySupervisedDC,appleShrivastavaPTSW16,m_Le-etal-ECCV18,Le_2019_CVPR_Workshops}. In this section, we show that we can modify shadow effects  using our proposed illumination model to generate additional training data. 

Given a shadow matte $\alpha$, a shadow-free image, and  parameters $(w,b)$, we can form a shadow image by:
\begin{equation}
    I^{\textrm{shadow}} = I^{\textrm{shadow-free}}\cdot \alpha + I^{\textrm{darkened}}\cdot (1-\alpha) \\
    \label{eq:aug}
\end{equation}

\noindent where $I^{\textrm{darkened}}$ has undergone the shadow effect associated to the set of shadow parameters $(w,b)$. Each pixel $i$ of $I^{\textrm{darkened}}$ is computed by:
\begin{equation}
I_i^{\textrm{darkened}} = (I_i^{\textrm{shadow-free}}-b) \cdot w^{-1}
\end{equation}

For each training image, we first compute the shadow parameters and the matte layer via Eqs. \ref{eq:trans} and \ref{eq:alpha}. 
Then, we  generate a new synthetic shadow image via Eq. \ref{eq:aug} with a scaling factor $w_{\textrm{syn}} = w \times k$. 
As  seen in Fig. \ref{fig:aug}, a lower $w$ leads to an image with a lighter shadow area while a higher $w$  increases the shadow effects instead. 
Using this method, we augment the ISTD training set by simply choosing $k=\left[0.8,0.9,1.1,1.2\right]$ to generate a new set of 5320 images, which is four times bigger than the original training set.
We augment the original ISTD dataset with this dataset.
 Training our model on this new augmented ISTD dataset improves our results, as the RMSE drops by 6\%, from 7.9 to 7.4, as reported in Tab. \ref{tab:all}. 

\def\subboxsize{0.3\subFigSzab}
\begin{figure}[]
 \centering
    \includegraphics[width=0.45\textwidth]{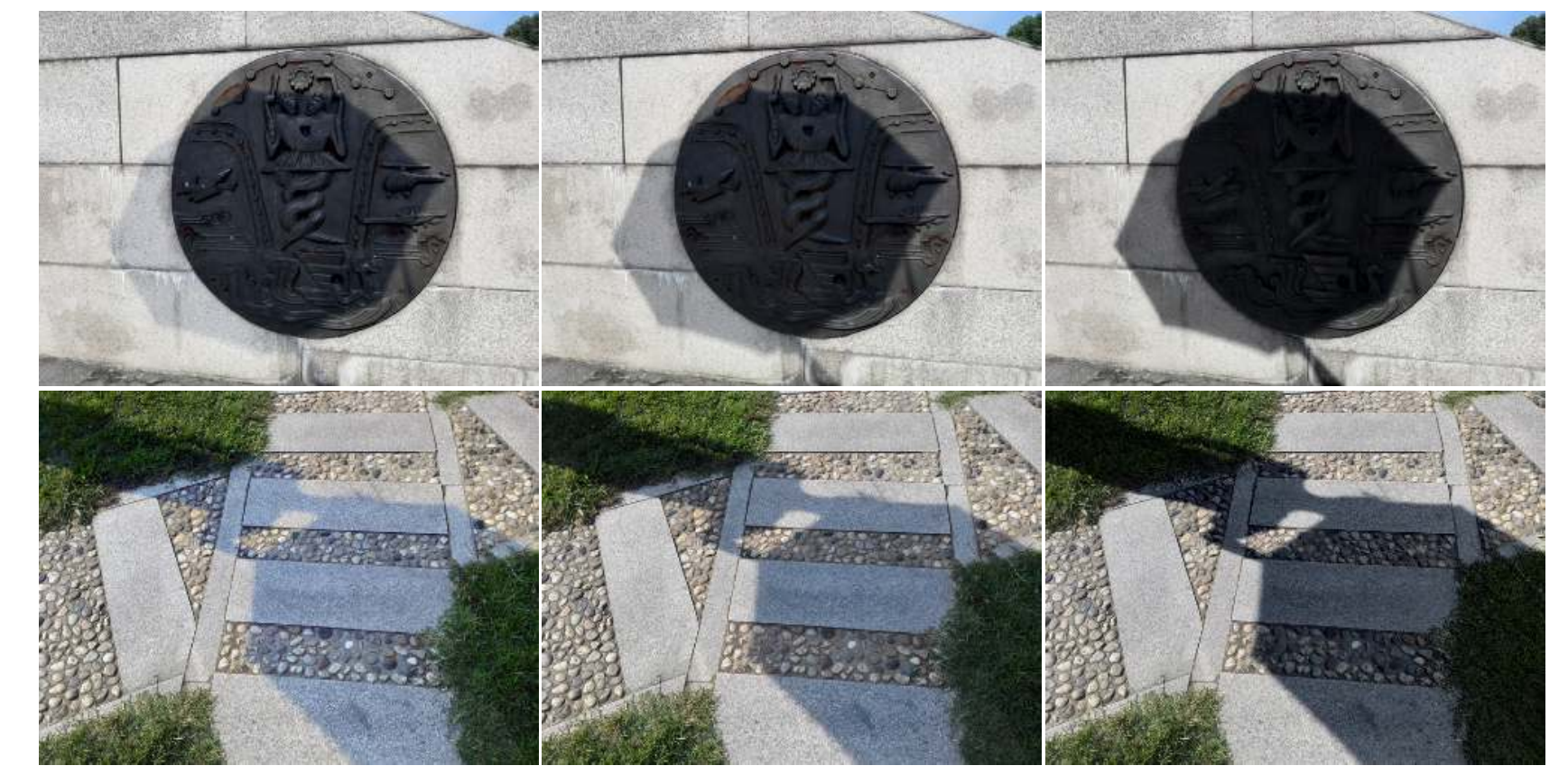}
     \makebox[\subboxsize]{Syns. Image}
    \makebox[\subboxsize]{Real Image}
    \makebox[\subboxsize]{Syns. Image}
        \makebox[\subboxsize]{ $w_{\textrm{syn}}=w\times 0.8$}
    \makebox[\subboxsize]{ }
    \makebox[\subboxsize]{$w_{\textrm{syn}}=w\times 1.7$}
    \caption{\textbf{Shadow editing via our decomposition model.} We use Eq. \ref{eq:aug} to generate synthetic shadow images. As we change the shadow parameters, the shadow effects change accordingly. 
    We show two example images from the ISTD training set where in the middle column are the original images and in the first and last column are synthetic.
    }
    \label{fig:aug}
\end{figure}

\begin{table}[]

\centering
\caption{\textbf{Shadow removal results of our networks train on the augmented ISTD dataset.} The metric is RMSE (the lower, the better).  Training our framework on the augumented ISTD dataset drops the RMSE on the shadow area from 7.9 to 7.4. }
\begin{tabular}{lcccc}
\toprule
Methods        &Train. Set            & Shad. & Non-Shad.& All  \\ 
\midrule
SP-Net &Aug. ISTD & 9.0  &3.2 &4.1\\ 
SP+M-Net &Aug. ISTD  & 7.4  &3.1 & 3.8\\
\midrule

\end{tabular}
\label{tab:all}
\end{table}



\def\subboxsize{0.12\subFigSzab}
\def\subfig{0.95\textwidth}
\begin{figure*}[]
 \centering

    \includegraphics[width=\subfig]{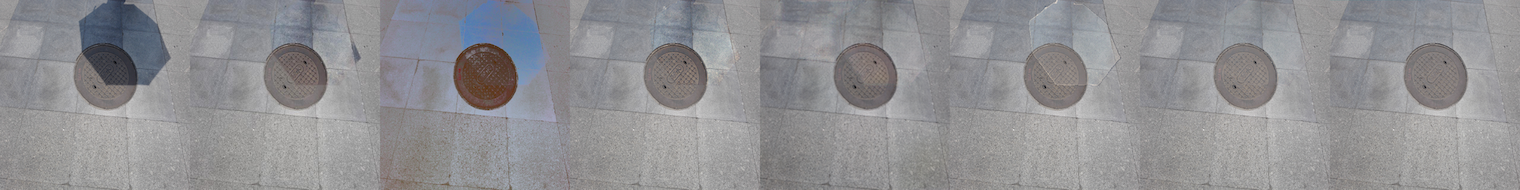}\\
    \includegraphics[width=\subfig]{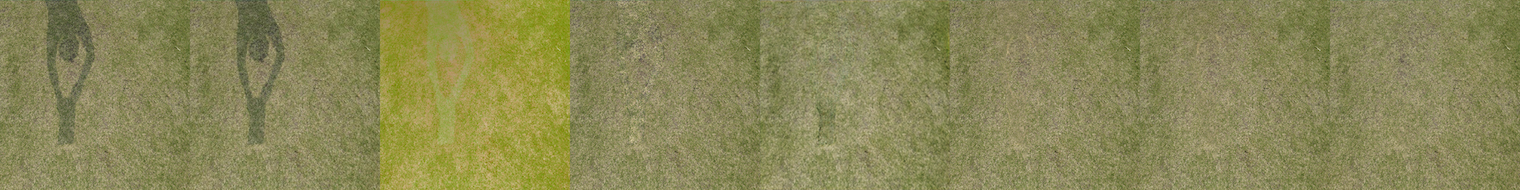}\\
    \includegraphics[width=\subfig]{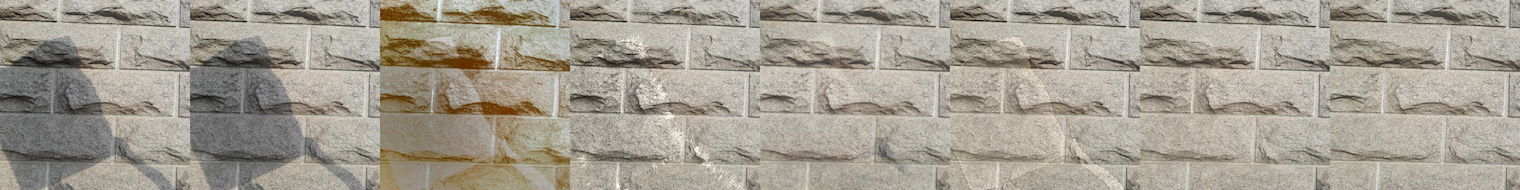}\\
    \includegraphics[width=\subfig]{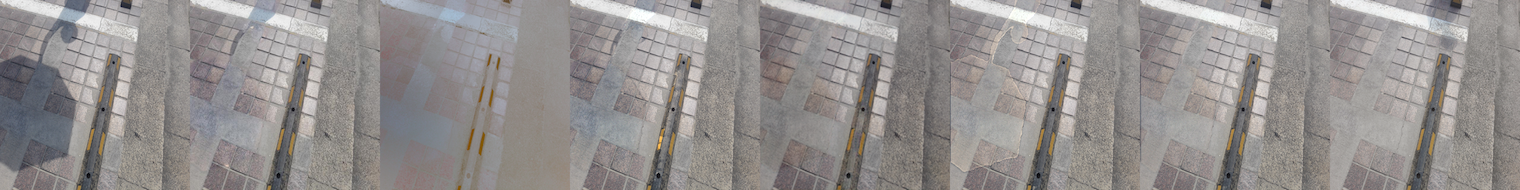}\\
        \includegraphics[width=\subfig]{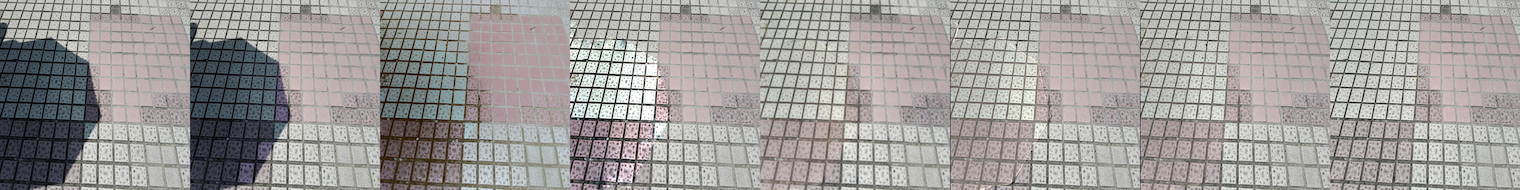}\\
     \includegraphics[width=\subfig]{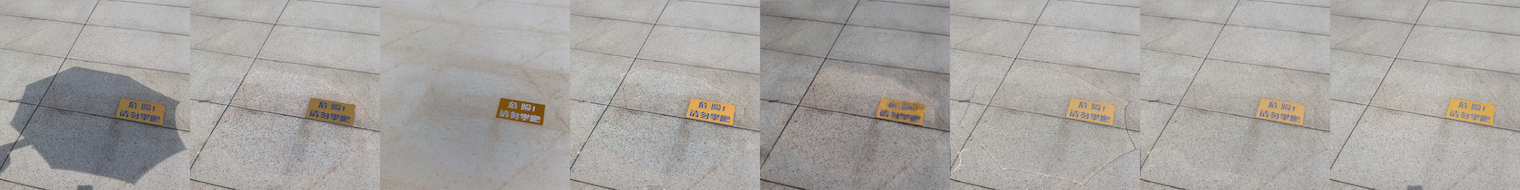}\\
    \makebox[\subboxsize]{Input}
    \makebox[\subboxsize]{Guo \etal}
    \makebox[\subboxsize]{Yang \etal}
    \makebox[\subboxsize]{Gong \etal}
    \makebox[\subboxsize]{Wang \etal }
    \makebox[\subboxsize]{SP-Net}
    \makebox[\subboxsize]{SP+M-Net}
    \makebox[\subboxsize]{Ground }\\
    \makebox[\subboxsize]{}
    \makebox[\subboxsize]{~\cite{guoPami}}
    \makebox[\subboxsize]{~\cite{Yang12}}
    \makebox[\subboxsize]{~\cite{Gong16}}
    \makebox[\subboxsize]{\cite{Wang_2018_CVPR}}
    \makebox[\subboxsize]{(Ours)}
    \makebox[\subboxsize]{(Ours)}
    \makebox[\subboxsize]{Truth}
  
     \caption{\textbf{Comparison of shadow removal on ISTD dataset.}     Qualitative comparison between our method and previous state-of-the-art methods: Guo \etal \cite{guoPami}, Yang \etal \cite{Yang12}, Gong \etal \cite{Gong16}, and Wang \etal \cite{Wang_2018_CVPR}. ``SP-Net'' are the shadow removal results using the parameters computed from SP-Net and a binary shadow mask. ``SP+M-Net'' are the  shadow removal results using the parameters computed from SP-Net and the shadow matte computed from M-Net.
    }
 
    \label{fig:main}
\end{figure*}

\section{Conclusions}
In this work, we have presented a novel framework for shadow removal in single images. Our main contribution is to use deep networks as the parameters estimators for an illumination model. 
Our approach has advantages over previous approaches. Comparing to the traditional methods using an illumination model for removing shadows, our deep networks can estimate the parameters for the model from a single image accurately and automatically. Comparing to deep learning methods that perform shadow removal via an end-to-end mapping, our shadow removal framework outputs images with high quality and no artifact since we do not use the deep network to output the per-pixel values.
Our model clearly achieves  state-of-the-art shadow removal results on the ISTD dataset. 
Our current approach can be extended in a number of ways. A more physically plausible illumination model would help the framework to output more realistic images. 
It would also be useful to develop a deep-learning based framework for shadow editing via a physical illumination model. 

\myheading{Acknowledgements.} 
This work was partially supported by the NSF EarthCube program (Award 1740595), the National Neographic/Microsoft AI for Earth program, the Partner University Fund, the SUNY2020 Infrastructure Transportation Security Center, and a gift from Adobe. Computational support provided by the Institute for Advanced Computational Science and a GPU donation from NVIDIA.
We thank Tomas Vicente for assistance with the manuscript.

\FloatBarrier
{\small
\bibliographystyle{ieee_fullname}
\bibliography{longstrings,egbib}
}


\end{document}